\documentclass[10pt, a4paper]{article}
\usepackage{lrec}
\usepackage{graphicx}
\usepackage{tabularx}
\usepackage{soul}
\usepackage[utf8]{inputenc}
\usepackage[export]{adjustbox}
\usepackage{epstopdf}	
\usepackage{comment}
\usepackage[breaklinks=true]{hyperref}
\expandafter\def\expandafter\UrlBreaks\expandafter{\UrlBreaks
  \do\a\do\b\do\c\do\d\do\e\do\f\do\g\do\h\do\i\do\j%
  \do\k\do\l\do\m\do\n\do\o\do\p\do\q\do\r\do\s\do\t%
  \do\u\do\v\do\w\do\x\do\y\do\z\do\A\do\B\do\C\do\D%
  \do\E\do\F\do\G\do\H\do\I\do\J\do\K\do\L\do\M\do\N%
  \do\O\do\P\do\Q\do\R\do\S\do\T\do\U\do\V\do\W\do\X%
  \do\Y\do\Z}
\usepackage{xstring}

\title{KGvec2go -- Knowledge Graph Embeddings as a Service}

\name{Jan Portisch (1,2), Michael Hladik (2), Heiko Paulheim (1)}

\address{(1) University of Mannheim - Data and Web Science Group, (2) SAP SE \\
         (1) B 6, 26 68159 Mannheim, Germany (2) Dietmar-Hopp Allee 16, 60190, Walldorf, Germany\\
         jan@informatik.uni-mannheim.de, michael.hladik@sap.com, heiko@informatik.uni-mannheim.de}

\abstract{
In this paper, we present KGvec2go, a Web API for accessing and consuming graph embeddings in a light-weight fashion in downstream applications.
Currently, we serve pre-trained embeddings for four knowledge graphs. 
We introduce the service and its usage, and we show further that the trained models have semantic value by evaluating them on multiple semantic benchmarks. 
The evaluation also reveals that the combination of multiple models can lead to a better outcome than the best individual model.\\ \newline  }

\begin{document}

\maketitleabstract

\section{Introduction}
A \textit{knowledge graph} (KG) stores factual information in the form of triples. Today, many such graphs exist for various domains, are publicly available, and are  being interlinked. As of 2019, the \textit{linked open data cloud} \cite{schmachtenberg2014adoption} counts more than 1,000 data sets with multiple billions of unique triples.\footnote{\url{https://lod-cloud.net/}} Knowledge graphs are typically consumed using factual queries for downstream tasks such as question answering. Recently, knowledge graph embedding models are explored as a new way of knowledge graph exploitation. \textit{KG embeddings} (KGEs) represent nodes and (depending on the approach) also edges as continuous vectors. One such approach is \textit{RDF2Vec} \cite{rdf_2_vec_2016}. 
It has been used and evaluated for machine learning, entity and document modeling, and for recommender systems \cite{rdf2vec_journal}. 
RDF2Vec vectors trained on a large knowledge graph have also been used as background knowledge source for ontology matching \cite{portisch_alod2vec_2018}.

While it has been shown that KGEs are helpful in many applications, embeddings on larger knowledge graphs can be expensive to train and to use for downstream applications. \url{kgvec2go.org}, therefore, allows to easily access and consume concept embeddings through simple Web APIs. 
Since most downstream applications only require embedding vectors for a small subset of all concepts, computing a complete embedding model or downloading a complete pre-computed one is often not desirable.

With \emph{KGvec2go}, rather than having to download the complete embedding model, a Web query can be used to obtain only the desired concept in vector representation or even a derived statistic such as the similarity between two concepts. This facilitates downstream applications on less powerful devices, such as smartphones, as well as the application of knowledge graph embeddings in machine learning scenarios where the data scientists do not want to train the models themselves or do not have the means to perform the computations.

Models for four knowledge graphs were learned, namely: 
\textit{DBpedia} \cite{dbpedia}, \textit{WebIsALOD} \cite{webisalod}, \textit{Wiktionary} \cite{dbnary_2015}, and \textit{WordNet} \cite{fellbaum_wordnet}.  \\
The data set presented here allows to compare the performance of different knowledge graph embeddings on different application tasks. It further allows to combine embeddings from different knowledge graphs in downstream applications.  We evaluated the embeddings on three semantic gold standards and also explored the combination of embeddings.\\
This paper is structured as follows: In the next section, related work will be presented. Section \ref{sec:approach} outlines the approach, Section \ref{sec:data_sets} presents the data sets for which an embedding has been trained, Section \ref{sec:api} introduces the Web API that is provided to consume the learned embedding models, and Section \ref{sec:evaluation} evaluates the models on three semantic gold standards. The paper closes with a summary and an outlook on future work.

\section{Related Work}
\label{sec:related_work}
For data mining applications, propositional feature vectors are required, i.e., vectors with either binary, nominal, or numerical elements. An RDF knowledge graph does not come with such properties and has to be translated into several feature vectors if it shall be exploited in data mining applications. This process is known as \textit{propositionalization} \cite{dzeroski_propositionalization_2001,ristoski2014comparison}. Two basic approaches for knowledge graph propositionalization can be distinguished: (i) Supervised propositionalization where the user has to manually craft features such as multiple ASK queries for nodes of interest and (ii) unsupervised approaches where the user does not have to know the structure of the graph. \cite{paulheim2012unsupervised}

In order to exploit knowledge graphs in data mining applications, embedding models have gained traction over the last years. \cite{wang2017knowledge} distinguish two families of approaches: distance based and semantic matching based approaches. The best known representatives of the first family are translation-based approaches. Given a set of entities $E$ and a set of edges $L$ as well as triples in the form $(head, label, tail)$, usually stated as $(h, l, t)$ where $h, t \in E$ and $l \in L$, \textit{TransE} \cite{transE}
trains vectors with the learning objective $h + l = t$ given that $(h,l,t)$ holds. Many similar approaches based on TransE have been proposed such as \textit{TransH} \cite{transH} or \textit{TransA} \cite{transA}.
In the second family, the most well known approaches are \emph{RESCAL} \cite{nickel2011three}, \emph{DistMult} \cite{yang2014embedding}, and \emph{HolE} \cite{nickel2016holographic}.

Another group of approaches exploits language models such as \textit{node2vec} \cite{node2vec} and \textit{RDF2Vec} \cite{rdf2vec_journal}. This work is based on the latter algorithm. 
Given a (knowledge) graph $G = (V,E)$ where $V$ is the set of vertices and $E$ is the set of directed edges, the RDF2Vec approach generates multiple sentences per vertex $v \in V$. An RDF2Vec sentence resembles a walk through the graph starting at a specified vertex $v$. Datatype properties are excluded from the walk generation. After the sentence generation, the \textit{word2vec} algorithm \cite{word_2_vec_1,word_2_vec_2} is applied to train a vector representation for each element $v \in V$ and $e \in E$.  word2vec is a neural language model. Given the context $k$ of a word $w$, where $k$ is a set of preceding and succeeding words of $w$, the learning objective of word2vec is to predict $w$. This is known as \textit{continuous bag of words} model (CBOW). The \textit{skip-gram} (SG) model is trained the other way around: Given $w$, $k$ has to be predicted. Within this training process, $c$ defines the size of $k$ and is also known as \textit{window} or \textit{window size}.\\
RDF2Vec is different from a pure language model in that it uses a knowledge graph as training corpus. Knowledge graphs are typically more structured than human language and can contain named entities that do not have to be explicitly detected. 

While there is an ever-growing number of knowledge graph embeddings, few works have addressed the software infrastructure aspect so far. The \emph{OpenKE} toolkit \cite{han2018openke} facilitates a unified framework for efficiently \emph{training} KGEs, but does not address the light-weight exploitation. The closest project to our work is \emph{WEmbedder} \cite{nielsen2017wembedder}, which, however, only serves embeddings for one single KG, i.e., Wikidata. This makes KGvec2go the first resource serving multiple embedding models simultaneously. 

\section{Approach}
\label{sec:approach}
For this work, the RDF2Vec approach has been re-implemented in Java and Python with a more efficient walk generation process. The implementation of the walk generator is publicly available on GitHub\footnote{\url{https://github.com/janothan/kgvec2go-walks/}}.\\
For the sentence generation, duplicate free random walks with \textit{depth = 8} have been generated  whereat edges within the sentences are also counted. For \textit{WordNet} and \textit{Wiktionary}, 500 walks have been calculated per entity. For \textit{WebIsALOD} and \textit{DBpedia}, 100 walks have been created in order to account for the comparatively large size of the knowledge graphs. 
\\   
The models were trained with the following configuration: \textit{skip-gram vectors}, \textit{window size = 5}, \textit{number of iterations = 5}, \textit{negative sampling for optimization}, \textit{negative samples = 25}. Apart from walk-generation adaptations due to the size of the knowledge graphs, the configuration parameters to train the models have been held constant and no data set specific optimizations have been performed in order to allow for comparability.\\
In addition, a Web API is provided to access the data models in a lightweight way. This allows for easy access to embedding models and to bring powerful embedding models to devices with restrictions in CPU and RAM, such as smart phones. The APIs are introduced in Section \ref{sec:api} The server has been implemented in Python using \textit{flask}\footnote{\url{https://flask.palletsprojects.com/en/1.1.x/}} and \textit{gensim} \cite{rehurek_lrec} and can be run using \textit{Apache HTTP Server}. Its code is publicly available on GitHub.\footnote{\url{https://github.com/janothan/kgvec2go-server/}}

\section{The Data Sets}
\label{sec:data_sets}
For this work, four data sets have been embedded which are quickly introduced in the following. 


\subsection{DBnary/Wiktionary}
\textit{Wiktionary} is "[a] collaborative project run by the Wikimedia Foundation to produce a free and complete dictionary in every language"\footnote{\url{https://web.archive.org/web/20190806080601/https://en.wiktionary.org/wiki/Wiktionary}}. The project is organized similarly to Wikipedia: Everybody can contribute and edit the dictionary. The content is reviewed in a community process. Like Wikipedia, Wiktionary is available in many languages. \textit{DBnary} \cite{dbnary_2015} is an RDF version of Wiktionary that is publicly available\footnote{\url{http://kaiko.getalp.org/about-dbnary/download/}}. The DBnary data set makes use of an extended \textit{LEMON} model \cite{mccrae_interchanging_2012} to describe the data. For this work, a recent download from July 2019 of the English Wiktionary has been used. 

\subsection{DBpedia}
\textit{DBpedia} is a well-known linked data set created by extracting structured knowledge from Wikipedia and other Wikimedia projects. The data is publicly available. For this work, the 2016-10 download has been used.\footnote{\url{https://wiki.dbpedia.org/downloads-2016-10}} Compared to the other knowledge graphs exploited here, DBpedia contains mainly instances such as the industrial rock band \textit{Nine Inch Nails} (which cannot be found in WordNet or Wiktionary). Therefore, DBpedia is with its instance data complementary to the other, lemma-focused, knowledge graphs.

\subsection{WebIsALOD} 
The \textit{WebIsA} database \cite{webisa_db} is a data set which consists of hypernymy relations extracted from the \textit{Common Crawl}\footnote{\url{https://commoncrawl.org/}}, a downloadable copy of the Web. The extraction was performed in an automatic manner through Hearst-like lexico-syntactic patterns. For example, from the sentence "[...] added that the country has favourable economic agreements with major economic powers, including the European Union.", the fact \texttt{isA(european\_union, major\_economic\_power)} is extracted\footnote{This is a real example, see: \url{http://webisa.webdatacommons.org/417880315}}.\\
\textit{WebIsALOD} \cite{webisalod} is the Linked Open Data endpoint which allows to query the data in SPARQL.\footnote{\url{http://webisa.webdatacommons.org/}} In addition to the endpoint, machine learning was used to assign confidence scores to the extracted triples. The data set of the endpoint is filtered, i.e. it contains a subset of the original WebIsA database, to ensure a higher data quality. The knowledge graph contains instances (like DBpedia) as well as more abstract concepts that can also be found in a dictionary.

\subsection{WordNet}
\textit{WordNet} \cite{fellbaum_wordnet} is a well-known and heavily used database of English word that are grouped in sets which represent one particular meaning, so-called \textit{synsets}. The resource is strictly authored. WordNet is publicly available, included in many natural language processing frameworks, and often used in research. An RDF version of the framework is also available for download and was used for this work.\footnote{\url{http://wordnet-rdf.princeton.edu/about/}}

\section{API}
\label{sec:api}
\url{kgvec2go.org} offers a simple Web API to retrieve: (i) individual vectors for concepts in different data sets, (ii) the cosine similarity between concepts directly, and (iii) the top $n$ most related concepts for any given concept. Alternatively, the full models can be downloaded from the Web site directly.\footnote{\url{http://www.kgvec2go.org/download.html}} The API is accessed through HTTP GET calls and will provide answers in the form of a JSON string. This allows for a simple usage on any device that has Internet access. In addition, natural words can be used to access the data rather than long URIs that follow their own idiosyncratic pattern as it is common for RDF2Vec embedded models. In the following, we will quickly describe the services that are offered. For a full description of the services as well as a graphical user interface to explore the embeddings, we refer to the Web page \url{kgvec2go.org}.

\subsection{Get Vector} 
\url{kgvec2go.org} allows to download an individual vector, i.e. a 200 dimensional floating point number array representation of a concept on a particular data set. The HTTP GET call follows the pattern below:   
\texttt{/rest/get-vector/<data\_set>/\\/<concept\_name>}\\
where \texttt{data\_set} refers to the data set that shall be used (i.e. one of \texttt{alod}, \texttt{dbpedia}, \texttt{wiktionary}, \texttt{wordnet}) and \texttt{concept\_name} to the natural language identifier of the concept (e.g. \textit{bed}). This call can be used in machine learning scenarios, for instance, where a numerical representation of a concept is required. \\
For data sets that learn an embedding based on the part-of-speech (POS) of the term, such as WordNet, multiple vectors are returned for one key word if the latter is available in multiple POS such as \textit{laugh} which occurs as noun and as verb.

\subsection{Get Similarity}
\begin{figure}
    \centering
    \includegraphics[scale=0.4, frame]{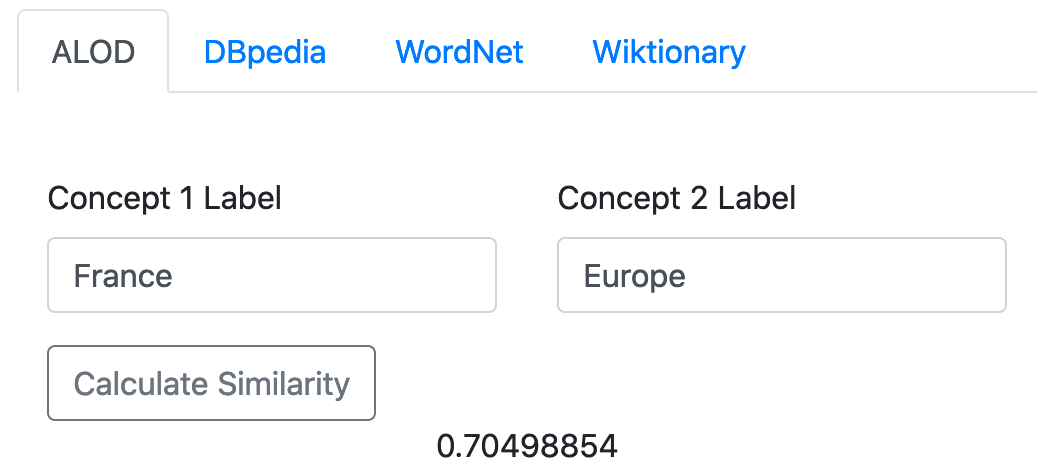}
    \caption{UI to query the similarity of two concepts online. Depicted is the similarity between \textit{France} and \textit{Europe} using the WebIsALOD embeddings.}
    \label{fig:similarity_ui}
\end{figure}

Given two concepts, \url{kgvec2go.org} allows to query a specified data set for the similarity score $s \in [-1.0,1.0]$ where $1.0$ refers to perfect similarity. 
The HTTP GET call follows the pattern below:  
\texttt{/rest/get-similarity/<data\_set>/\\<concept\_name\_1>/<concept\_name\_2>}\\
where \texttt{data\_set} refers to the set that shall be used and the two concept names refer to the concept labels for which the similarity shall be calculated. This call can be used wherever the similarity or relatedness of two concepts needs to be judged such as in recommender systems or matching tasks. A Web UI is available to try out this call in a Web browser.\footnote{\url{http://www.kgvec2go.org/query.html}} A screenshot is shown in Figure \ref{fig:similarity_ui} for the terms \textit{France} and \textit{Europe} for the model learned on WebIsALOD.

{}

\subsection{Get Closest Concepts}
\begin{figure}
\centering
\includegraphics[scale=0.35, frame]{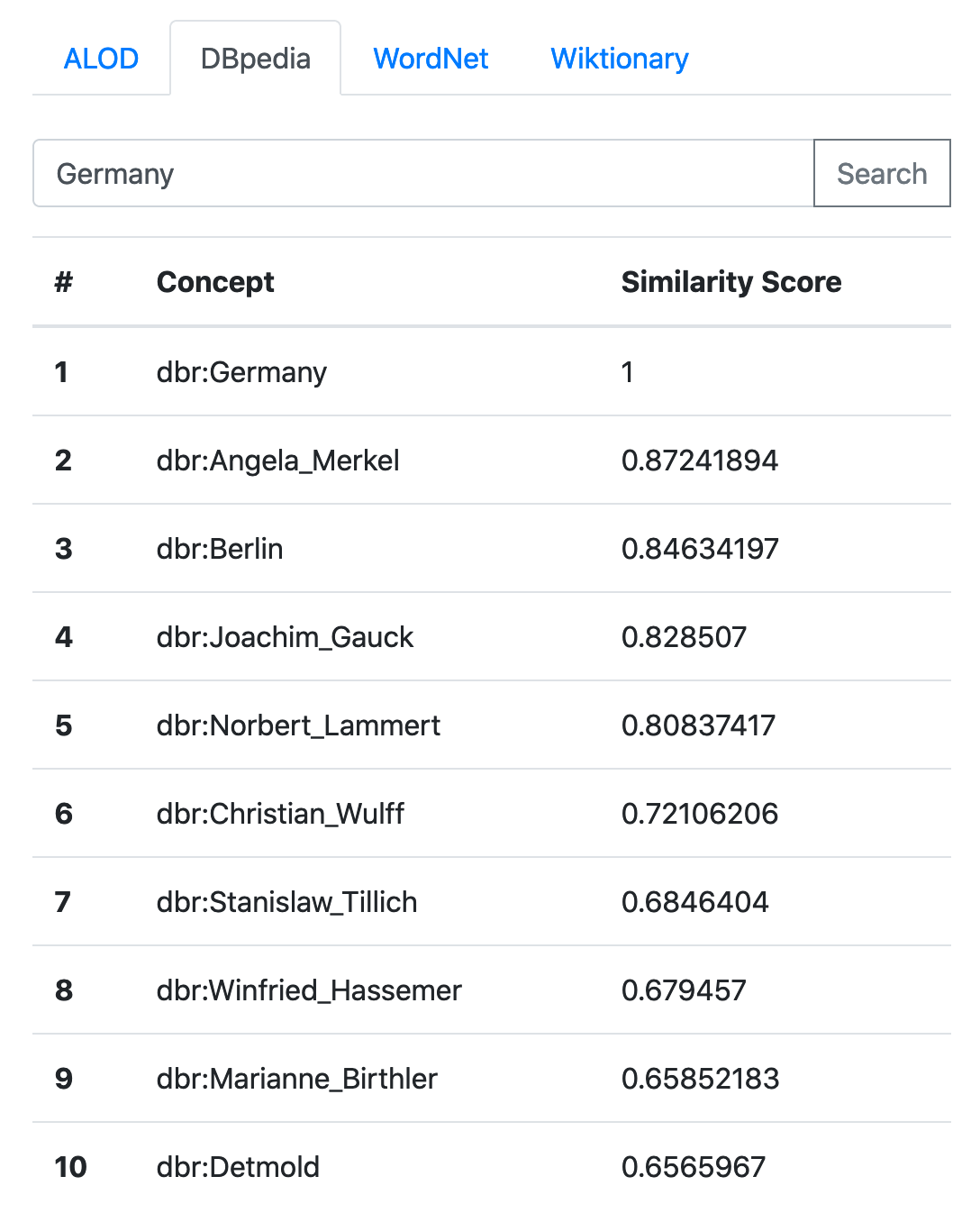}
\caption{UI to query the data set online. Shown is the result for query term \textit{Germany} on data set DBpedia. Note that the underlying DBpedia version for the training is that of 2016. In that year, Angela Merkel was the Chancellor of Germany, Berlin the capital of the country, Joachim Gauck the President of Germany, and Norbert Lammert the President of the Bundestag.}
\label{fig:closest_query_ui}
\end{figure}

The API is also capable of determining the closest $n$ concepts given a concept and a data set. The given concept is mapped to the vector space and compared with all other vectors. Therefore, the call is expensive on large data sets and should rather be used to explore the data set. 
The HTTP GET call follows the pattern below:  
\texttt{/rest/closest-concepts/<data\_set>/\\<top\_n>/<concept\_name>}\\
where \texttt{data\_set} refers to the set that shall be used, \texttt{top\_n} refers to the number of closest concepts that shall be obtained, and \texttt{concept\_name} refers to the written representation of the concept. For data sets that learn an embedding based on the part-of-speech of the term, such as WordNet, all closest concepts are determined for all POS of the term and their scores are summarized. This allows to calculate the $n$ closest concepts for a single term, such as \textit{sleep}, that occurs in multiple POS (in this case as noun and as verb).\\
A Web UI is available to try out this call in a Web browser.\footnote{\url{http://www.kgvec2go.org/query.html}} A screenshot is shown in Figure \ref{fig:closest_query_ui} for the term \textit{Germany} on the trained DBpedia model.

\section{Evaluation} 
\label{sec:evaluation}

\subsection{Evaluation Gold Standards}
In order to test whether there is semantic value in the trained vectors, we evaluate them on three data sets: \textit{WordSim-353} \cite{wordsim}, \textit{SimLex-999} \cite{simlex}, and \textit{MEN} \cite{men-gs}. The principle of evaluation is the same for all gold standards used: The system is presented with two words and has to determine their relatedness or similarity; then, the rank correlation (also known as \textit{Spearman's Rho}) with the scores in the gold standards is calculated. Higher correlations between the gold standards' scores and the system's scores are regarded as better. Pairs with an out of vocabulary term are handled here by returning a similarity of $0$. As the goal of this data set are comparable general purpose embeddings, it is important to note that the embeddings were not specifically trained to perform well on the given tasks. On similarity tasks, for instance, the results would likely improve when antonymy relations were dropped. With other configuration settings, it is also possible to improve the results further on the given evaluation sets; this has, for instance, been done in \cite{madoc52029} where better relatedness/similarity results on WebIsALOD could be achieved with other RDF2Vec configurations.

\subsection{Evaluation Mode}
\label{ssec:evaluation_mode}
The learned models were evaluated on their own on each of the evaluation data sets. In addition, a combination of all data sets was evaluated. Therefore, the individual similarity scores were added. Hence, $s_{combined}(c_1, c_2) = s_{DBpedia}(c_1, c_2) + s_{WebIsALOD}(c_1, c_2) +s_{Wiktionary}(c_1, c_2) + s_{WordNet}(c_1, c_2)$ where $s_{combined}$ is the final similarity score assigned to the concept pair $c_1$ and $c_2$ and $s_{data set}$ describes the individual score of a model trained on a single \textit{data set} for the same concept pair. This can be done without normalization because (i) all scores are in the same value range ($[-1,1]$), (ii) out of vocabulary terms receive a score of 0 (so they do not influence the final results), and (iii) because Spearman's rank correlation is used which is independent of the absolute values -- only the rank is considered.

\subsection{Evaluation Results}
The rank correlations on the three gold standards are summarized in Table \ref{tab:evaluation-results}. It can be seen that the results vary depending on the gold standard used. The Wiktionary data set performs best when it comes to relatedness. The WebIsALOD data set performs similarly well on WS-353 and performs best on MEN. On the SimLex-999 gold standard, WordNet outperforms the other data sets. The performance of DBpedia is significantly worse which is due to many out of vocabulary terms: This particular data set is focused on instance data rather than lexical forms such as \textit{angry}. The evaluation performed here is, therefore, not optimal for the data set. This can also be observed in the example results depicted in Table~\ref{tab:examples}: While DBpedia and WebIsALOD work well for entities such as \emph{Germany}, Wiktionary performs better for general words such as \textit{loud}.

Interestingly, the combined evaluation mode outlined in subsection \ref{ssec:evaluation_mode} is able to outperform the best individual results on WS-353 ($\rho = 0.678$ vs. $\rho = 0.571$) as well as on MEN ($\rho = 0.230$ vs. $\rho = 0.207$). On SimLex, the combination of all similarity scores is very close to the best individual score (WordNet). This shows that it can be beneficial to combine several embedding spaces on different data sets.\\
It is important to note that the vectors were not trained for the specific task at hand. Nonetheless, the combined embeddings perform well on WS-353 albeit top-notch systems for each data set cannot be outperformed. By the lower performance on SimLex-999 and MEN it can be seen that relatedness is better represented in the embedding spaces than actual similarity. This is an intuitive result given that there was no training objective towards similarity.\\
When looking at the different properties of the knowledge graphs, it can be reasoned that the level of authoring is not important for the performance on the tasks at hand: Web\-IsALOD embeddings, which are derived from an automatically generated knowledge graph, easily outperform WordNet embeddings, which are derived from a highly authored knowledge base, on WS-353 and MEN.

\subsection{Further Remarks}
It is also possible to find typical analogies in the data. In this case, two concepts are presented to the model together with a third one for which the system shall determine an analogous concept. In the following examples, the underlined concept is the best concept that the system found given the three non-underlined concepts.\\

For example, on Wiktionary:
\begin{itemize}
  \item \textit{girl} is to \textit{boy} like \textit{man} is to \underline{\textit{woman}} 
  \item \textit{big} is to \textit{small} like \textit{fake} is to \underline{\textit{original}}
  \item \textit{beautiful} is to \textit{attractive} like \textit{quick} is to \underline{\textit{rapid}}
\end{itemize}
Similar results can be found on instance level. For example, on DBpedia:
\begin{itemize}
    \item \textit{Germany} is to \textit{Angela Merkel} like \textit{France} is to \underline{\textit{François Hollande}}\footnote{Note that François Hollande is indeed the president of France as of 2016.}
\end{itemize}


\begin{table}[]
\begin{tabular}{|l|l|l|l|}
\hline
                   & \textbf{WS-353} & \textbf{SimLex-999} & \textbf{MEN} \\ \hline
\textbf{Wiktionary}& 0.5708         & 0.2265             & 0.1513\\ \hline
\textbf{DBpedia}   & 0.1430         & -0.0097            & 0.0804  \\ \hline
\textbf{WebIsALOD} & 0.5598         & 0.1509             & 0.2066 \\ \hline
\textbf{WordNet}   & 0.4074         & 0.2870             & 0.1086 \\ \hline
\textbf{Combined}  & 0.6784         & 0.2815             & 0.2304 \\ \hline
\end{tabular}
\caption{Rank correlation of the data sets with three gold standards.}
\label{tab:evaluation-results}
\end{table}

\begin{table*}[t]
    \centering
    \begin{tabular}{r|l|l|l|l}
        & Wiktionary & DBpedia & WebIsALOD & WordNet \\
        \hline
        1 & Germany & Germany & europe & Germany \\
        2 & snazziness & Angela Merkel & uk & FRG \\
        3 & West Germany & Berlin & france & skillet \\
        4 & these islands & Joachim Gauck & canada & Federal Republic of Germany \\
        5 & cobbler & Norbert Lammert & japan & Deutschland \\
        6 & German Empire & Christian Wulff & italy & High German \\
        7 & derisive & Stanislaw Tillich & australia & German \\
        8 & who shot John & Winfried Hassemer & usa & Pietism \\
        9 & glute & Marianne Birthler & england & Bavaria \\
        10 & Okla. & Detmold & asia & ingrained \\
        \hline
        1 & loud & Loud & cons fan & loud (s)\\
        2 & silent & Loli & scream & secondly \\
        3 & noiseless & Cometa (HVDC) & weird noise & loud (r)\\
        4 & rackety & Looc & of noise & aright \\
        5 & noisy & Loob & history of 20th century & loud (a)\\
        6 & unsilent & Python Server Pages & collective sigh of relief & fruticulose \\
        7 & piercing & Louk & thwack & red-handed \\
        8 & quiet & Juan Llort & undesired signal & deep down \\
        9 & clamorous & Lojo & grinning & every bit \\
        10 & blasting & Lone & complaint of office worker & rhymeless \\
        \hline
    \end{tabular}
    \caption{Example results for the search terms \emph{Germany} (upper part) and \emph{loud} (lower part). WordNet returns \textit{loud} multiple times with different part-of-speech tags. On DBpedia, results for \textit{Loud} are given as there is no vector for \textit{loud}.}
    \label{tab:examples}
\end{table*}

\section{Summary and Future Work}
\label{sec:summary_future_work}
In this paper, we presented \texttt{KGvec2go}, a resource consisting of trained embedding models on four knowledge graphs. The models were evaluated on three different gold standards. It could be shown, that the trained vectors carry semantic meaning and that a combination of different knowledge graph embeddings can be beneficial in some tasks. Furthermore, a lightweight API was presented which allows to consume the models in a computationally cheap, memory-efficient, and easy way through Web APIs. We are confident that our work eases the usage of knowledge graph embeddings in real-world applications.

For the future, we plan to extend the data set by adding more different embedding models of knowledge graphs to the resource presented, as well as including other knowledge graphs, and to extend the capabilities of the current API. Furthermore, we plan to exploit the trained models for downstream application tasks that profit from the inclusion of background knowledge such as ontology matching and domain specific data integration tasks.

\section{Bibliographical References}
\label{main:ref}

\bibliographystyle{lrec}
\bibliography{references}


\end{document}